\documentclass[11pt,a4paper]{article}
\usepackage[hyperref]{acl2020}
\usepackage{times}
\usepackage{latexsym}
\usepackage{todonotes}

\usepackage{algorithm2e}
\usepackage{amsmath}
\usepackage{url}
\usepackage{hyperref}
\usepackage{url}
\usepackage{graphicx}
\usepackage{float}
\usepackage{amssymb}
\usepackage{enumitem}
\usepackage{multirow}
\usepackage{hyperref}       
\usepackage{subcaption}

\SetKwRepeat{Do}{do}{while}%

\usepackage[symbol]{footmisc}

\usepackage{url}
\usepackage{listings}

\usepackage{booktabs, tabularx}

\aclfinalcopy 

\renewcommand{\thefootnote}{\fnsymbol{footnote}}

\title{BPE-Dropout: Simple and Effective Subword Regularization}

  \author{Ivan Provilkov\footnotemark[1]$^{\ \ 1,2}$ \quad Dmitrii Emelianenko\footnotemark[1]$^{\ \ 1,3}$ \quad Elena Voita$^{4,5}$\bigskip\\
  $^1$Yandex, Russia \\
  $^2$Moscow Institute of Physics and Technology, Russia\\
  $^3$National Research University Higher School of Economics, Russia\\
  $^4$University of Edinburgh, Scotland
  \quad $^5$University of Amsterdam, Netherlands\\
  {\tt \{iv-provilkov, dimdi-y, lena-voita\}@yandex-team.ru}
}

\date{}

\begin{document}
\maketitle
\begin{abstract}

  Subword segmentation is widely used to address the open vocabulary problem in machine translation. The dominant approach to subword segmentation is Byte Pair Encoding~(BPE), which keeps the most frequent words intact while splitting the rare ones into multiple tokens. While multiple segmentations are possible even with the same vocabulary, BPE splits words into unique sequences; this may prevent a model from better learning the compositionality of words and being robust to segmentation errors. 
So far, the only way to overcome this BPE imperfection, its deterministic nature, was to create another subword segmentation algorithm~\cite{sentencepiece}. In contrast, we show that BPE itself incorporates the ability to produce multiple segmentations of the same word. 
We introduce \textit{BPE-dropout} -- simple and effective subword regularization method based on and compatible with conventional BPE.
It stochastically corrupts the segmentation procedure of BPE, which leads to producing multiple segmentations within the same fixed BPE framework.
Using \textit{BPE-dropout} during training and the standard BPE during inference improves translation quality up to 2{.}3 BLEU compared to BPE and up to 0{.}9 BLEU compared to the previous subword regularization.
\footnotetext[1]{Equal contribution.}

\end{abstract}

\renewcommand{\thefootnote}{\arabic{footnote}}

\section{Introduction}

Using subword segmentation has become de-facto standard in Neural Machine Translation~\cite{bojar-etal-2018-findings,barrault-etal-2019-findings}. Byte Pair Encoding~(BPE)~\cite{sennrich-etal-2016-neural} is the dominant approach to subword segmentation. It keeps the common words intact while splitting the rare and unknown ones into a sequence of subword units. This potentially allows a model to make use of morphology, word composition and transliteration. BPE effectively deals with an open-vocabulary problem and is widely used due to its simplicity.

\begin{figure*}[t!]
    \centering
    \begin{subfigure}[b]{0.18\textwidth}
        \includegraphics[width=\textwidth]{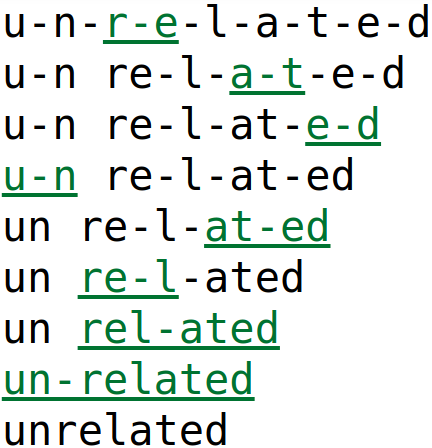}
        \caption{}
        \label{fig:examples_original}
    \end{subfigure}
    \quad\quad\quad\quad
    \begin{subfigure}[b]{0.65\textwidth}
        \includegraphics[width=\textwidth]{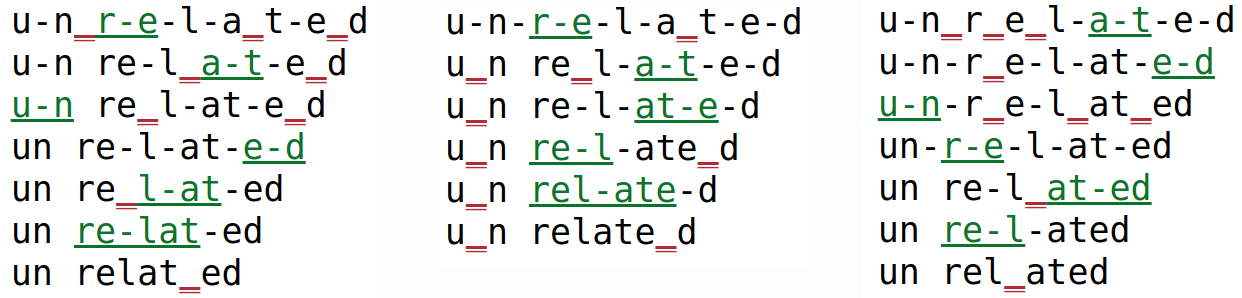}
        \caption{}
        \label{fig:examples_dropout}
    \end{subfigure}
    \vspace{-1ex}
    \caption{Segmentation process of the word `\textit{unrelated}' using (a) BPE, (b) \textit{BPE-dropout}. Hyphens indicate possible merges (merges which are present in the merge table); merges performed at each iteration are shown in green, dropped -- in red. }\label{fig:intro_examples}
    \vspace{-2ex}
\end{figure*}

There is, however, a drawback of BPE in its deterministic nature: it splits words into unique subword sequences, which means that for each word a model observes only one segmentation. Thus, a model is likely not to reach its full potential in exploiting morphology, learning the compositionality of words and being robust to segmentation errors. Moreover, as we will show further, subwords into which rare words are segmented end up poorly understood.

A natural way to handle this problem is to enable multiple segmentation candidates. This was initially proposed by \citet{sentencepiece} as a subword regularization -- a regularization method, which is implemented as an on-the-fly data sampling and is not specific to NMT architecture.
Since standard BPE produces single segmentation, to realize this regularization the author had to propose a new subword segmentation, different from BPE. However, the introduced approach is rather complicated: it requires training a separate segmentation unigram language model, using EM and Viterbi algorithms, and forbids using conventional BPE.

In contrast, we show that BPE itself incorporates the ability to produce multiple segmentations of the same word. BPE builds a vocabulary of subwords and a merge table, which specifies which subwords have to be merged into a bigger subword, as well as the priority of the merges. During segmentation, words are first split into sequences of characters, then the learned merge operations are applied to merge the characters into larger, known symbols, till no merge can be done (Figure~\ref{fig:intro_examples}(a)). We introduce \textit{BPE-dropout} -- a subword regularization method based on and compatible with conventional BPE. It uses a vocabulary and a merge table built by BPE, but at each merge step, some merges are randomly dropped. This results in different segmentations for the same word~(Figure~\ref{fig:intro_examples}(b)). 
Our method requires no segmentation training in addition to BPE and uses standard BPE at test time, therefore is simple. \textit{BPE-dropout} is superior compared to both BPE and~\citet{sentencepiece} on a wide range of translation tasks, therefore is effective.

Our key contributions are as follows:
\begin{itemize}
    \item We introduce \textit{BPE-dropout} -- a simple and effective subword regularization method;
    \item We show that our method outperforms both BPE and previous subword regularization on a wide range of translation tasks;
    \item We analyze how training with \textit{BPE-dropout}  affects a model and show that it leads to a better quality of learned token embeddings and to a model being more robust to noisy input.
\end{itemize}

\section{Background}

In this section, we briefly describe BPE and the concept of subword regularization.
We assume that our task is machine translation, where a model needs to predict the target sentence $Y$ given the source sentence $X$, but the methods we describe are not task-specific.

\subsection{Byte Pair Encoding (BPE)}\label{sec:bpe}

To define a segmentation procedure, BPE~\cite{sennrich-etal-2016-neural} builds a token vocabulary and a merge table.
The token vocabulary is initialized with the character vocabulary, and the merge table is initialized with an empty table. First, each word is represented as a sequence of tokens plus a special end of word symbol. Then, the method iteratively counts all pairs of tokens and merges the most frequent pair into a new token. This token is added to the vocabulary, and the merge operation is added to the merge table. This is done until the desired vocabulary size is reached. 

The resulting merge table specifies which subwords have to be merged into a bigger subword, as well as the priority of the merges. In this way, it defines the segmentation procedure. First, a word is split into distinct characters plus the end of word symbol. Then, the pair of adjacent tokens which has the highest priority is merged. This is done iteratively until no merge from the table is available~(Figure~\ref{fig:intro_examples}(a)).

\subsection{Subword regularization}\label{sec:subword_reg}

Subword regularization~\cite{sentencepiece} is a training algorithm which integrates multiple segmentation candidates. Instead of maximizing log-likelihood, this algorithm maximizes log-likelihood marginalized over different segmentation candidates. Formally,
\vspace{-2ex}
\begin{equation} \label{eq:objective}
\begin{split}
\end{split}
\mathcal{L} = \sum_{(X,Y)\in D}{\mathop{\mathbb{E}}_{\substack{x \sim P(x | X) \\ y \sim P(y | Y)}} {\log P (y | x, \theta)}},
\end{equation}
where $ x $ and $ y $ are sampled segmentation candidates for sentences $ X $ and $ Y $ respectively, $P(x|X)$ and $P(y|Y)$ are the probability distributions the candidates are sampled from, and $ \theta $ is the set of model parameters. In practice, at each training step only one segmentation candidate is sampled.

Since standard BPE segmentation is deterministic, to realize this regularization~\citet{sentencepiece} proposed a new subword segmentation. The introduced approach requires training a separate segmentation unigram language model to predict the probability of each subword, EM algorithm to optimize the vocabulary, and Viterbi algorithm to make samples of segmentations.

Subword regularization was shown to achieve significant improvements over the method using a single subword sequence. However, the proposed method is rather complicated and forbids using conventional BPE. This may prevent practitioners from using subword regularization.

\section{Our Approach: BPE-Dropout}\label{sec:method}

We show that to realize subword regularization it is not necessary to reject BPE since multiple segmentation candidates can be generated within the BPE framework. We introduce \textit{BPE-dropout} -- a method which exploits the innate ability of BPE to be stochastic. It alters the segmentation procedure while keeping the original BPE merge table. During segmentation, at each merge step some merges are randomly dropped with the probability $p$. This procedure is described in Algorithm~1.

\vspace{-1ex}

\begin{algorithm}
 \vspace{5px}
 \hrule
 \vspace{3px}
 \textbf{Algorithm 1: BPE-dropout} \hrule
 \label{alg:bpedrop}
    $ current\_split  \leftarrow $ characters from input\_word\;
    \Do {
        $ merges $ is not empty
    } {
        $ merges  \leftarrow $ all possible merges\footnote{lll} of tokens from $ current\_split $\;
        \For{ $ merge $ from $ merges $} {
            \tcc{The only difference from BPE}
            remove $ merge $ from $ merges $ with the probability $ p $\; } 
        \If{ $ merges $ is not empty} {
            $ merge  \leftarrow $ select the merge with the highest priority from $ merges $\;
            apply $ merge $ to $ current\_split $\;
        }
    }
    \Return{current\_split}\;
 \hrule
 
\end{algorithm}

\footnotetext[1]{In case of multiple occurrences of the same merge in a word (for example, \texttt{m-e-r-g-e-r} has two occurrences of the merge (\texttt{e, r})), we decide independently for each occurrence whether to drop it or not.}

\vspace{-1ex}

If $p$ is set to 0, the segmentation is equivalent to the standard BPE; if $p$ is set to 1, the segmentation splits words into distinct characters. The values between 0 and 1 can be used to control the segmentation granularity. 

We use $p>0$ (usually $p=0{.}1$) in training time to expose a model to different segmentations and $p=0$ during inference, which means that at inference time we use the original BPE. We discuss the choice of the value of $p$ in Section~\ref{sec:experiments}.

When some merges are randomly forbidden during segmentation, words end up segmented in different subwords; see for example Figure~\ref{fig:intro_examples}(b). We hypothesize that exposing a model to different segmentations may result in better understanding of the whole words as well as their subword units; we will verify this in Section~\ref{sect:analysis}.

\section{Experimental setup}
\label{sec:exp_setup}

\subsection{Baselines}

Our baselines are the standard BPE and the subword regularization by~\citet{sentencepiece}.

Subword regularization by~\citet{sentencepiece} has segmentation sampling hyperparameters $l$ and $\alpha$. $l$ specifies how many best segmentations for each word are produced before sampling one of them, $\alpha$ controls the smoothness of the sampling distribution. In the original paper $(l=\infty, \alpha=0{.}2/0{.}5)$ and  $(l=64, \alpha=0{.}1)$ were shown to perform best on different datasets. Since overall they show comparable results, in all experiments we use $(l=64, \alpha=0{.}1)$.

\subsection{Vocabularies}

There are two ways of building vocabulary for models trained with \textit{BPE-dropout}:
(1) take the vocabulary built by BPE; then the segmented with \textit{BPE-dropout} text will contain a small number of unknown tokens (\textsc{UNK}s)\footnote{For example, for the English part of the IWSLT15 En-Vi corpora, these \textsc{UNK}s make up 0{.}00585 and 0{.}00085 of all tokens for 32k and 4k vocabularies, respectively.};
    (2) add to the BPE vocabulary all tokens which can appear when segmenting with \textit{BPE-dropout}. 

In the preliminary experiments, we did not observe any difference in quality; therefore, either of the methods can be used. We choose the first option to stay in the same setting as the standard BPE. Besides, a model exposed to some \textsc{UNK}s in training can be more reliable for practical applications where unknown tokens can be present.


\subsection{Data sets and preprocessing}

\begin{table*}[t!]
\centering
\begin{tabular}{lccccc}
\toprule
 & & Number of sentences & Voc size & Batch size & The value of $p$\\
 & & (train/dev/test) &  &  & in \textit{BPE-dropout}\\

\midrule
IWSLT15& En $\leftrightarrow$ Vi & 133k / 1553 / 1268 & 4k & 4k & 0.1 / 0.1\\
& En $\leftrightarrow$ Zh & 209k / 887 / 1261 &  4k / 16k  & 4k & 0.1 / 0.6\\
\cmidrule{2-6}
IWSLT17& En $\leftrightarrow$ Fr & 232k / 890 / 1210 & 4k & 4k & 0.1 / 0.1\\
& En $\leftrightarrow$ Ar & 231k / 888 / 1205 & 4k & 4k & 0.1 / 0.1\\
\cmidrule{2-6}
WMT14 & En $\leftrightarrow$ De & 4{.}5M / 3000 / 3003 & 32k & 32k  & 0.1 / 0.1\\
\cmidrule{2-6}
ASPEC & En $\leftrightarrow$ Ja & 2M / 1700 / 1812 & 16k & 32k & 0.1 / 0.6\\
\bottomrule
\end{tabular}
\caption{Overview of the datasets and dataset-dependent hyperparametes; values of $p$ are shown in pairs: source language / target language. (We explain the choice of the value of $p$ for \textit{BPE-dropout} in Section~\ref{sect:choice_of_p}.)}
\label{tab:datasets}
\end{table*}

We conduct our experiments on a wide range of datasets with different corpora sizes and languages; information about the datasets is summarized in Table~\ref{tab:datasets}. These datasets are used  in the main experiments~(Section~\ref{sect:main_results}) and were chosen to match the ones used in the prior work~\cite{sentencepiece}. In the additional experiments~(Sections~\ref{sect:drop_on_one_side}-\ref{sect:inference_time}), we also use random subsets of the WMT14 English-French data; in this case, we specify dataset size for each experiment.

Prior to segmentation, we preprocess all datasets with the standard Moses toolkit.\footnote{\url{https://github.com/moses-smt/mosesdecoder}}
However, Chinese and Japanese have no explicit word boundaries, and Moses tokenizer does not segment sentences into words; for these languages, subword segmentations are trained almost from unsegmented raw sentences.

Relying on a recent study of how the choice of vocabulary size influences translation quality~\cite{lowresBPE}, we choose vocabulary size depending on the dataset size (Table~\ref{tab:datasets}). 

In training, translation pairs  were  batched  together  by  approximate  sequence length. For the main experiments, the values of batch size we used are given in Table~\ref{tab:datasets} (batch size is the number of source tokens). In the experiments in Sections~\ref{sect:drop_on_one_side}, \ref{sect:choice_of_p} and \ref{sect:corpora_voc_size}, for datasets not larger than 500k sentence pairs we use vocabulary size and batch size of 4k, and 32k for the rest.\footnote{Large batch size can be reached by using several of GPUs or by accumulating the gradients for several batches and then making an update.}

In the main text, we train all models on lowercased data. In the appendix, we provide additional experiments with the original case and case-sensitive BLEU.

\subsection{Model and optimizer}
\label{sect:model_parameters}

The NMT system used in our experiments is {\it Transformer base}~\cite{transformer}.  More precisely, the number of layers is $N=6$ with $h = 8$ parallel attention layers, or heads. The dimensionality of input and output is $d_{model} = 512$, and the inner-layer of feed-forward networks has dimensionality $d_{ff}=2048$. We use regularization and optimization procedure as described in~\citet{transformer}.

\subsection{Training time}

We train models till convergence. For all experiments, we provide number of training batches  in the appendix (Tables~\ref{tab:num_steps_main} and \ref{tab:num_steps_one_side}).

\subsection{Inference}

To produce translations, for all models,  we use beam search with the beam of 4 and length normalization of 0.6.

In addition to the main results, \citet{sentencepiece} also report scores using $n$-best decoding. To translate a sentence, this strategy produces multiple segmentations of a source sentence, generates a translation for each of them, and rescores the obtained translations. While this could be an interesting future work to investigate different sampling and rescoring strategies, in the current study we use 1-best decoding to fit in the standard decoding paradigm.

\subsection{Evaluation}
For evaluation, we average 5 latest checkpoints and use BLEU~\cite{BLEU} computed via SacreBleu\footnote{Our SacreBLEU signature is: $\texttt{BLEU+case.lc+}$ $\texttt{lang.[src-lang]-[dst-lang]+numrefs.1+}$ $\texttt{smooth.exp+tok.13a+version.1.3.6}$} \cite{sacrebleu}. For Chinese, we add option $\texttt{--tok zh}$ to SacreBLEU. For Japanese, we use character-based BLEU. 


\section{Experiments}\label{sec:experiments}

\subsection{Main results}
\label{sect:main_results}

\begin{table}[t!]
\centering
\begin{tabular}{lcccc}
\toprule

 & & BPE  & \citet{sentencepiece} & \textit{BPE-dropout}\\

\midrule
\multicolumn{5}{l}{\!\!\!\bf IWSLT15}\\
&\!\!\!\!\!\! En-Vi & 31{.}78 & 32{.}43 & \bf{33{.}27} \\
&\!\!\!\!\!\! Vi-En & 30{.}83 & 32{.}36 & \bf{32{.}99} \\
&\!\!\!\!\!\! En-Zh & 20.48 & \bf{23.01}  & \bf{22.84} \\
&\!\!\!\!\!\! Zh-En & 19.72 & 21.10 & \bf{21.45} \\
\midrule
\multicolumn{5}{l}{\!\!\!\bf IWSLT17}\\
&\!\!\!\!\!\! En-Fr & 39{.}37 &	39{.}45 &	\bf{40{.}02} \\
&\!\!\!\!\!\! Fr-En & 38{.}18 &	38{.}88 &	\bf{39{.}39} \\
&\!\!\!\!\!\! En-Ar & 13.89 & 14.43 & \bf{15.05} \\
&\!\!\!\!\!\! Ar-En & 31.90 & 32.80 & \bf{33.72} \\
\midrule
\multicolumn{5}{l}{\!\!\!\bf WMT14}\\
&\!\!\!\!\!\! En-De & 27.41 & \bf{27.82} & \bf{28.01}  \\
&\!\!\!\!\!\! De-En & 32.69 & 33.65 & \bf{34.19}\\
\midrule
\multicolumn{5}{l}{\!\!\!\bf ASPEC}\\
&\!\!\!\!\!\! En-Ja & 54.51 & \bf{55.46} & 55.00 \\
&\!\!\!\!\!\! Ja-En & 30.77 & \bf{31.23} & \bf{31.29} \\

\bottomrule
\end{tabular}
\caption{BLEU scores. Bold indicates the best score and all scores whose difference from the best is not statistically significant (with $p$-value of 0.05). (Statistical significance is computed via bootstrapping~\cite{koehn2004statistical}.)}
\label{tab:main}
\end{table}

The results are provided in Table~\ref{tab:main}.
For all datasets, \textit{BPE-dropout} improves
significantly over the standard BPE: more than 1{.}5 BLEU for En-Vi, Vi-En, En-Zh, Zh-En, Ar-En, De-En,  and 0{.}5-1{.}4 BLEU for the rest. The improvements are especially prominent for smaller datasets; we will discuss this further in Section~\ref{sect:corpora_voc_size}.

Compared to \citet{sentencepiece}, among the 12 datasets we use \textit{BPE-dropout} is beneficial for 8 datasets with improvements up to  0{.}92 BLEU, is not significantly different for 3 datasets and underperforms only on En-Ja. While \citet{sentencepiece} uses another segmentation, our method operates within the BPE framework and changes only the way a model is trained. Thus, lower performance of \textit{BPE-dropout} on En-Ja and only small or insignificant differences for Ja-En, En-Zh and Zh-En suggest that Japanese and Chinese may benefit from a language-specific segmentation.

Note also that \citet{sentencepiece} report larger improvements over BPE from using their method than we show in Table~\ref{tab:main}. This might be explained by the fact that \citet{sentencepiece} used large vocabulary size (16k, 32k), which has been shown counterproductive for small datasets~\cite{sennrich-zhang-2019-revisiting,lowresBPE}. While this may not be the issue for models trained with subword regularization~(see Section~\ref{sect:corpora_voc_size}), this causes drastic drop in performance of the baselines.

\begin{table}[t!]
\centering
\begin{tabular}{lcccc}
\toprule
 & BPE & \multicolumn{3}{c}{\textit{BPE-dropout}}\\
 & & src-only  & dst-only & both\\

\midrule
250k & 26{.}94 & 27{.}98 & 27{.}71 & \bf{28{.}40} \\
500k & 29{.}28 & \bf{30{.}12} & 29{.}40 & \bf{29{.}89} \\
1m & 30{.}53 & \bf{31{.}09} & 30{.}62 & \bf{31{.}23} \\
4m & 33{.}38 & \bf{33{.}89} & 33{.}46 & \bf{33{.}85} \\
16m & 34{.}37 & \bf{34{.}82} & - & 33{.}66 \\
\bottomrule
\end{tabular}
\caption{BLEU scores for models trained with \textit{BPE-dropout} on a single side of a translation pair or on both sides. Models trained on random subsets of WMT14 En-Fr dataset. Bold indicates the best score and all scores whose difference from the best is not statistically significant (with $p$-value of 0.05).}
\label{tab:one_side}
\end{table}

\subsection{Single side vs full regularization}
\label{sect:drop_on_one_side}

In this section, we investigate whether \textit{BPE-dropout} should be used only on one side of a translation pair or for both source and target languages. We select random subsets of different sizes from WMT14 En-Fr data to understand how the results are affected by the amount of data. We show that:
\begin{itemize}
    \item for small and medium datasets, full regularization performs best;
    \item for large datasets, \textit{BPE-dropout} should be used only on the source side.
\end{itemize}

Since full regularization performs the best for most of the considered dataset sizes, in the subsequent sections we use \textit{BPE-dropout} on both source and target sides.

\subsubsection{Small and medium datasets: use full regularization}

Table~\ref{tab:one_side} indicates that using \textit{BPE-dropout} on the source side is more beneficial than on the target side; for the datasets not smaller than 0.5m sentence pairs, \textit{BPE-dropout} can be used only the source side. We can speculate that it is more important for the model to understand a source sentence than being exposed to different ways to generate the same target sentence.

\subsubsection{Large datasets: use only for source}

For larger corpora (e.g., starting from 4m instances),  it is better to use \textit{BPE-dropout} only on the source side (Table~\ref{tab:one_side}). Interestingly, using \textit{BPE-dropout} for both source and target languages hurts performance for large datasets.

\subsection{Choice of the value of $p$}
\label{sect:choice_of_p}

\begin{figure}[t!]
    \centering
    \includegraphics[width=5cm]{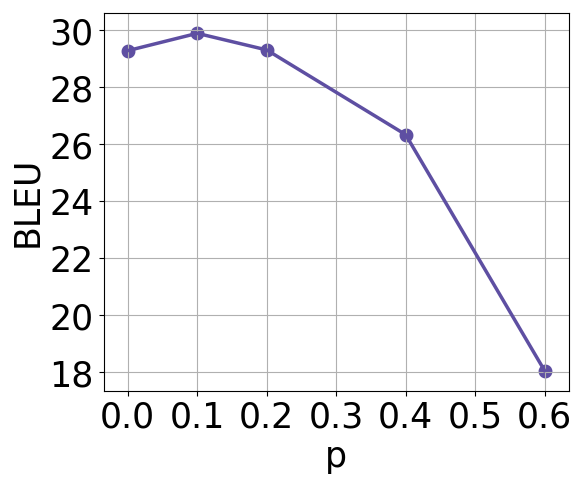}
    \caption{BLEU scores for the models trained with \textit{BPE-dropout} with different values of $ p $. WMT14 En-Fr, 500k sentence pairs.}
    \label{fig:drop_grid}
\end{figure}

Figure~\ref{fig:drop_grid} shows BLEU scores for the models trained on \textit{BPE-dropout} with different values of $ p $ (the probability of a merge being dropped). 
Models trained with high values of $p$ are unable to translate due to a large mismatch between training segmentation (which is close to char-level) and inference segmentation (BPE). The best quality is achieved with $p = 0{.}1$. 

In our experiments, we use $p = 0{.}1$ for all languages except for Chinese and Japanese.
For Chinese and Japanese, we take the value of $p=0.6$ to match the increase in length of segmented sentences for other languages.\footnote{Formally, for English/French/etc. with \textit{BPE-dropout}, $p = 0.1$ sentences become on average about 1.25 times longer compared to segmented with BPE; for Chinese and Japanese, we need to set the value of $p$ to $0{.}6$ to achieve the same increase.}

\subsection{Varying corpora and vocabulary size}

Now we will look more closely at how the improvement from using \textit{BPE-dropout} depends on corpora and vocabulary size. 

\begin{figure}[t!]
    \centering
    \includegraphics[width=7.5cm]{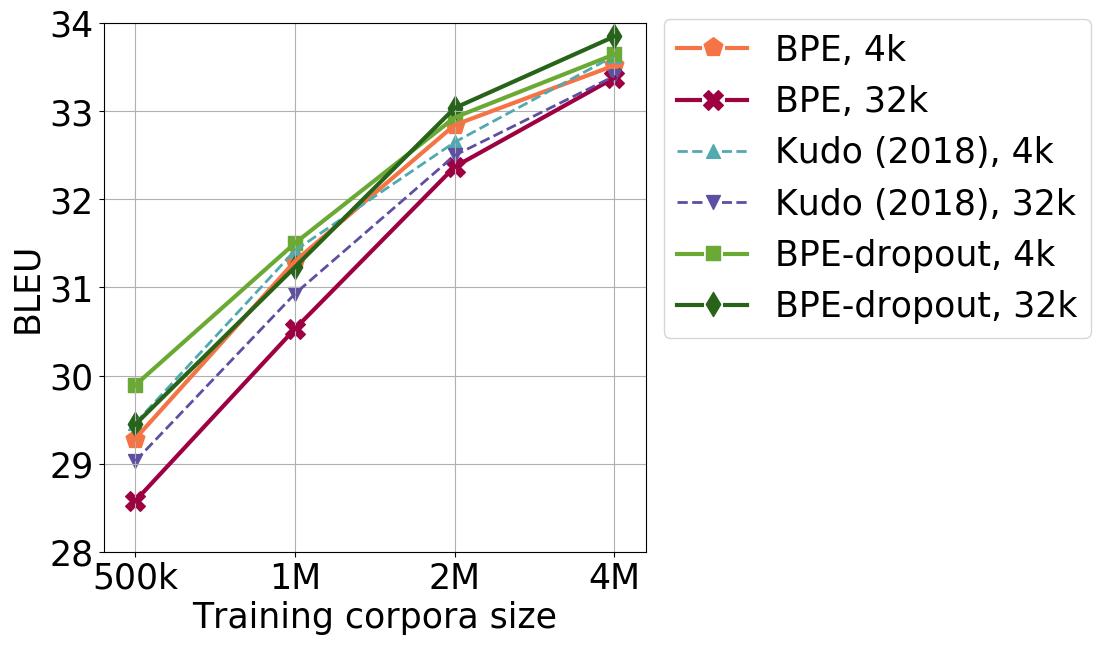}
    \caption{BLEU scores. Models trained on random subsets of WMT14 En-Fr.}
    \label{fig:corpora_size}
\end{figure}

First, we see that \textit{BPE-dropout} performs best for all dataset sizes~(Figure~\ref{fig:corpora_size}). Next, models trained with subword regularization are less sensitive to the choice of vocabulary size: differences in performance of models with 4k and 32k vocabulary are much less than for models trained with the standard BPE. This makes \textit{BPE-dropout} attractive since it allows (i)~not to tune vocabulary size for each dataset, (ii)~choose vocabulary size depending on the desired model properties: models with smaller vocabularies are beneficial in terms of number of parameters, models with larger vocabularies are beneficial in terms of inference time.\footnote{Table~\ref{tab:timing} shows that inference for models with 4k vocabulary is more than 1{.}4 times longer than models with 32k vocabulary.} Finally, we see that the effect from using \textit{BPE-dropout}  vanishes when a corpora size gets bigger. This is not surprising: the effect of any regularization is less in high-resource settings; however, as we will show later in Section~\ref{sect:misspel}, when applied to noisy source, models trained with \textit{BPE-dropout} show substantial improvements up to 2 BLEU even in high-resource settings. 

Note that for larger corpora, we recommend using \textit{BPE-dropout} only for source language~(Section~\ref{sect:drop_on_one_side}).

\label{sect:corpora_voc_size}

\subsection{Inference time and length of generated sequences}
\label{sect:inference_time}

\begin{figure}[t!]
    \centering
    \begin{subfigure}[b]{0.23\textwidth}
        \includegraphics[width=\textwidth]{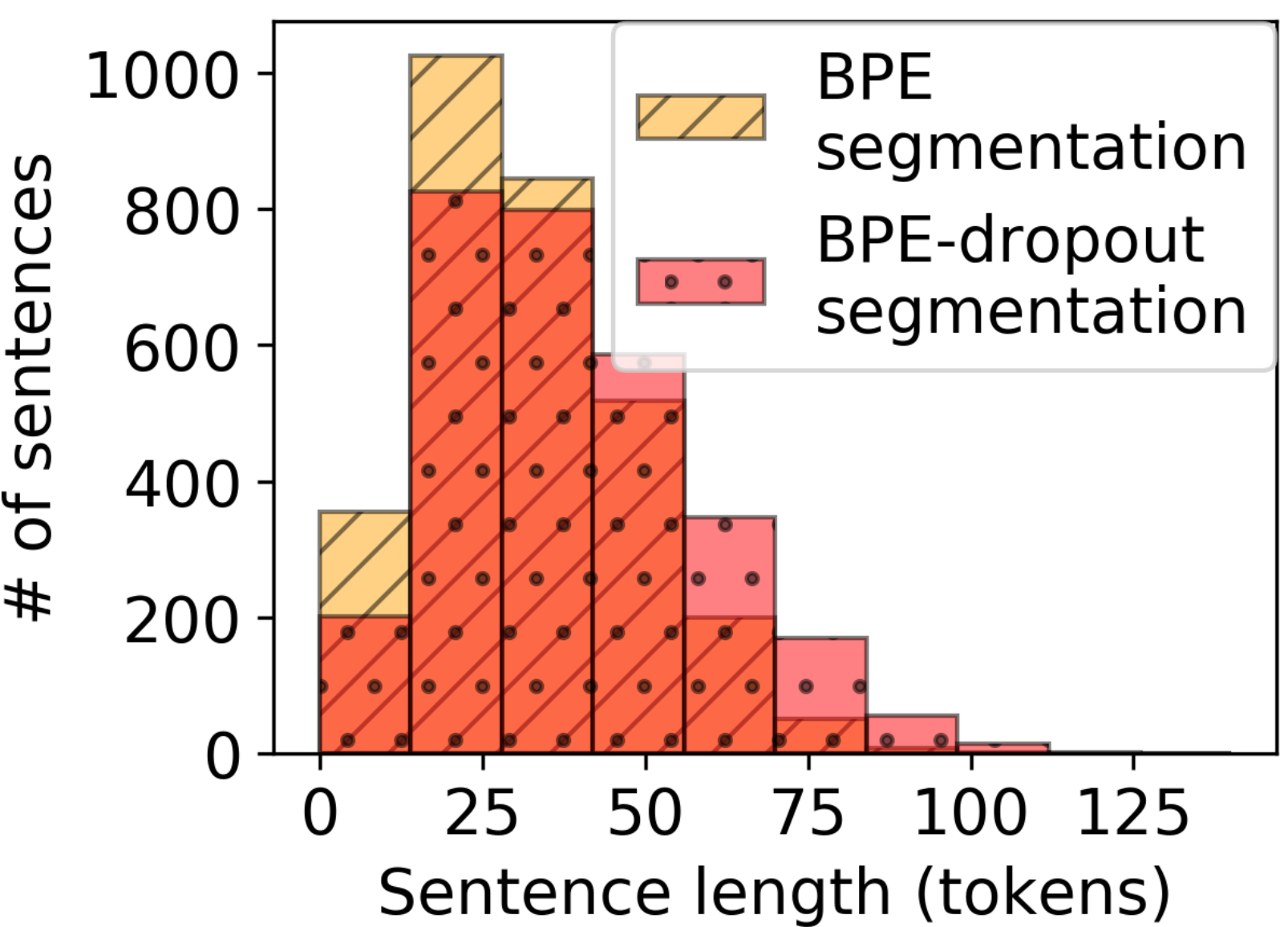}
        \caption{}
    \end{subfigure}
    \begin{subfigure}[b]{0.23\textwidth}
       \includegraphics[width=\textwidth]{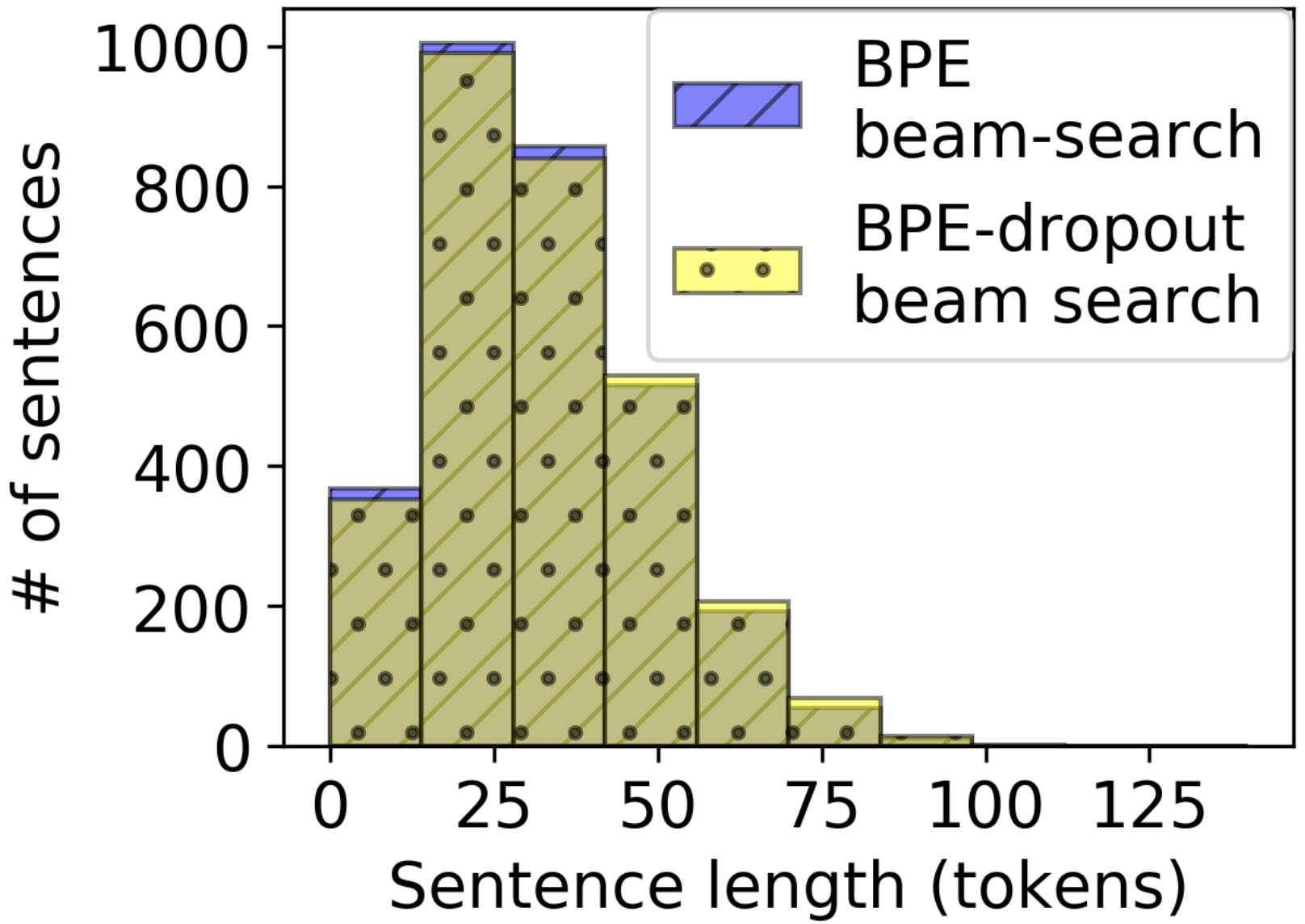}
        \caption{}
    \end{subfigure}
    \caption{Distributions of length~(in tokens) of (a) the French part of WMT14 En-Fr test set segmented using BPE or  \textit{BPE-dropout}; and (b) the generated translations for the same test set by models trained with BPE or \textit{BPE-dropout}.}
    \label{fig:output_length}
\end{figure}

Since \textit{BPE-dropout} produces more fine-grained segmentation, sentences segmented with \textit{BPE-dropout} are longer; distribution of sentence lengths are shown in Figure~\ref{fig:output_length}~(a) (with $p=0{.}1$,  on average about 1{.}25 times longer). Thus there is a potential danger that models trained with \textit{BPE-dropout} may tend to use more fine-grained segmentation in inference and hence to slow inference down. However, in practice this is not the case: distributions of lengths of generated translations for models trained with BPE and with \textit{BPE-dropout} are close (Figure~\ref{fig:output_length}~(b)).\footnote{This is the result of using beam search: while samples from a model reproduce training data distribution quite well, beam search favors more frequent tokens~\cite{ott2018analyzing}. Therefore, beam search translations tend not to use less frequent fine-grained segmentation.}

Table~\ref{tab:timing} confirms these observations and shows that inference time of models trained with \textit{BPE-dropout} is not substantially different from the ones trained with BPE.

\begin{table}[t!]
\centering
\begin{tabular}{l|cc}
\toprule
voc size & BPE & \textit{BPE-dropout}\\
\midrule
32k & 1.0 & 1.03 \\
4k & 1.44 & 1.46 \\

\bottomrule
\end{tabular}
\caption{Relative inference time of models trained with different subword segmentation methods. Results obtained by (1) computing averaged over 1000 runs time needed to translate WMT14 En-Fr test set, (2) dividing all results by the smallest of the obtained times. }
\label{tab:timing}
\end{table}

\section{Analysis}
\label{sect:analysis}

In this section, we analyze qualitative differences between models trained with BPE and \textit{BPE-dropout}. We find, that 
\begin{itemize}
    \item when using BPE, frequent sequences of characters rarely appear in a segmented text as individual tokens rather than being a part bigger ones; \textit{BPE-dropout} alleviates this issue;
    \item by analyzing the learned embedding spaces, we show that using \textit{BPE-dropout} leads to a better understanding of rare tokens;
    \item as a consequence of the above, models trained with \textit{BPE-dropout} are more robust to misspelled input.
\end{itemize}

\begin{figure}[t!]
    \centering
    
    \includegraphics[scale=0.4]{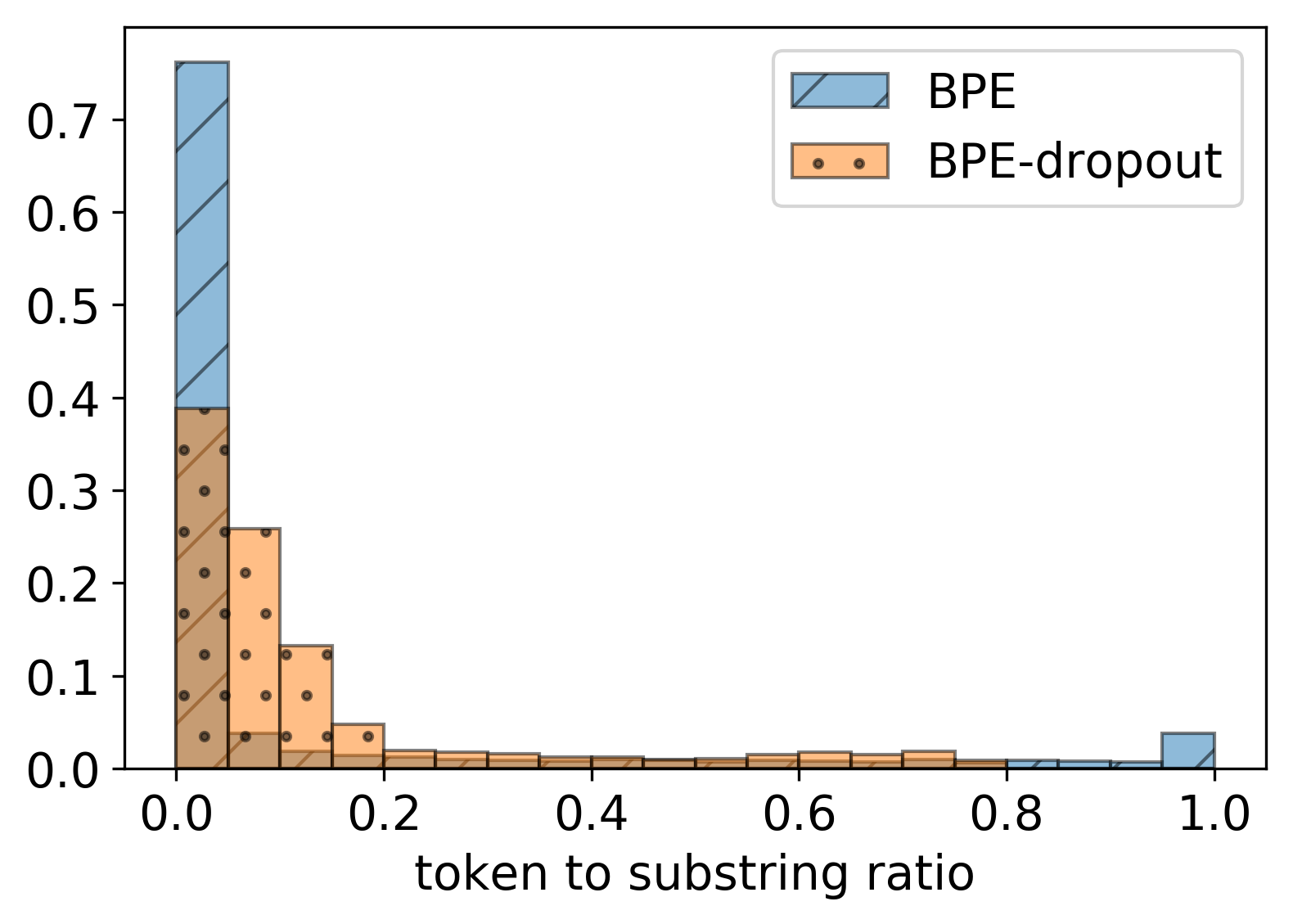}

    \caption{Distribution of token to substring ratio for texts segmented using BPE or \textit{BPE-dropout} for the same vocabulary of 32k tokens; only 10$\%$ most frequent substrings are shown.  (Token to substring ratio of a token is the ratio between its frequency as an individual token and as a sequence of characters.)  }\label{fig:token_frequency}
\end{figure}

\begin{figure*}[t!]
    \centering
    \includegraphics[scale=0.27]{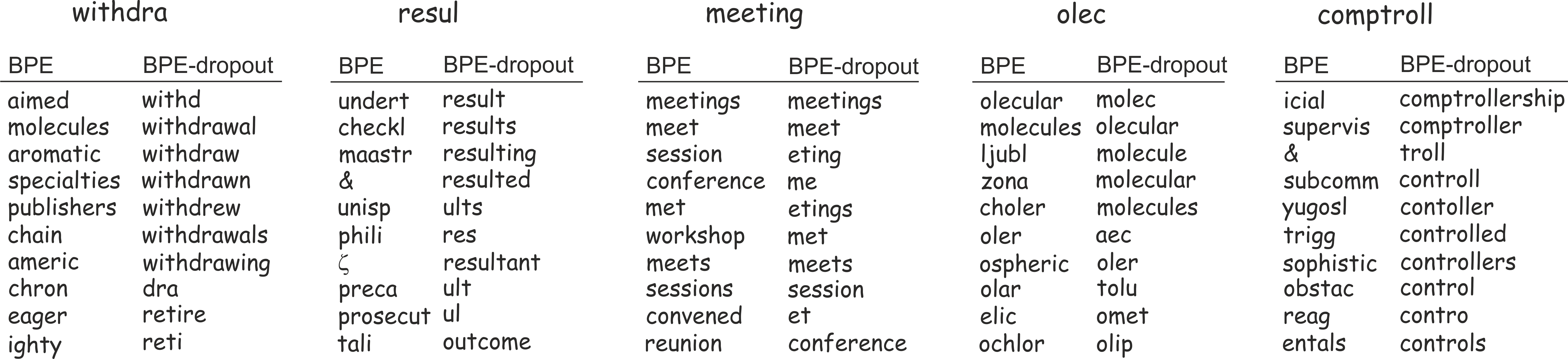}
    \caption{Examples of nearest neighbours in the source embedding space of models trained with BPE and \textit{BPE-dropout}. Models trained on WMT14 En-Fr (4m).}
    \label{fig:closest_embs}
\end{figure*} 

\subsection{Substring frequency}

Here we highlight one of the drawbacks of BPE's deterministic nature: since it splits words into unique subword sequences, only rare words are split into subwords. This forces frequent sequences of characters to mostly appear in a segmented text as part of bigger tokens, and not as individual tokens. To show this, for each token in the BPE vocabulary we calculate how often it appears in a segmented text as an individual token and as a sequence of characters (which may be part of a bigger token or an individual token). Figure~\ref{fig:token_frequency} shows distribution of the ratio between substring frequency as an individual token and as a sequence of characters (for top-10$\%$ most frequent substrings).

For frequent substrings, the distribution of token to substring ratio is clearly shifted to zero, which confirms our hypothesis: frequent sequences of characters rarely appear in a segmented text as individual tokens. When a text is segmented using \textit{BPE-dropout} with the same vocabulary, this distribution significantly shifts away from zero, meaning that frequent substrings appear in a segmented text as individual tokens more often.

\subsection{Properties of the learned embeddings}

Now we will analyze embedding spaces learned by different models.
We take embeddings learned by models trained with BPE and \textit{BPE-dropout} and for each token look at the closest neighbors in the corresponding embedding space.
Figure~\ref{fig:closest_embs} shows several examples. In contrast to BPE, nearest neighbours of a token in the embedding space of \textit{BPE-dropout} are often tokens that share sequences of characters with the original token. To verify this observation quantitatively, we computed character 4-gram precision of top-10 neighbors: the proportion of those 4-grams of the top-10 closest neighbors which are present among 4-grams of the original token. As expected, embeddings of \textit{BPE-dropout} have higher character 4-gram precision~(0.29) compared to the precision of BPE~(0.18).

\begin{figure}[t!]
    \centering
    \begin{subfigure}[b]{0.22\textwidth}
        \includegraphics[width=\textwidth]{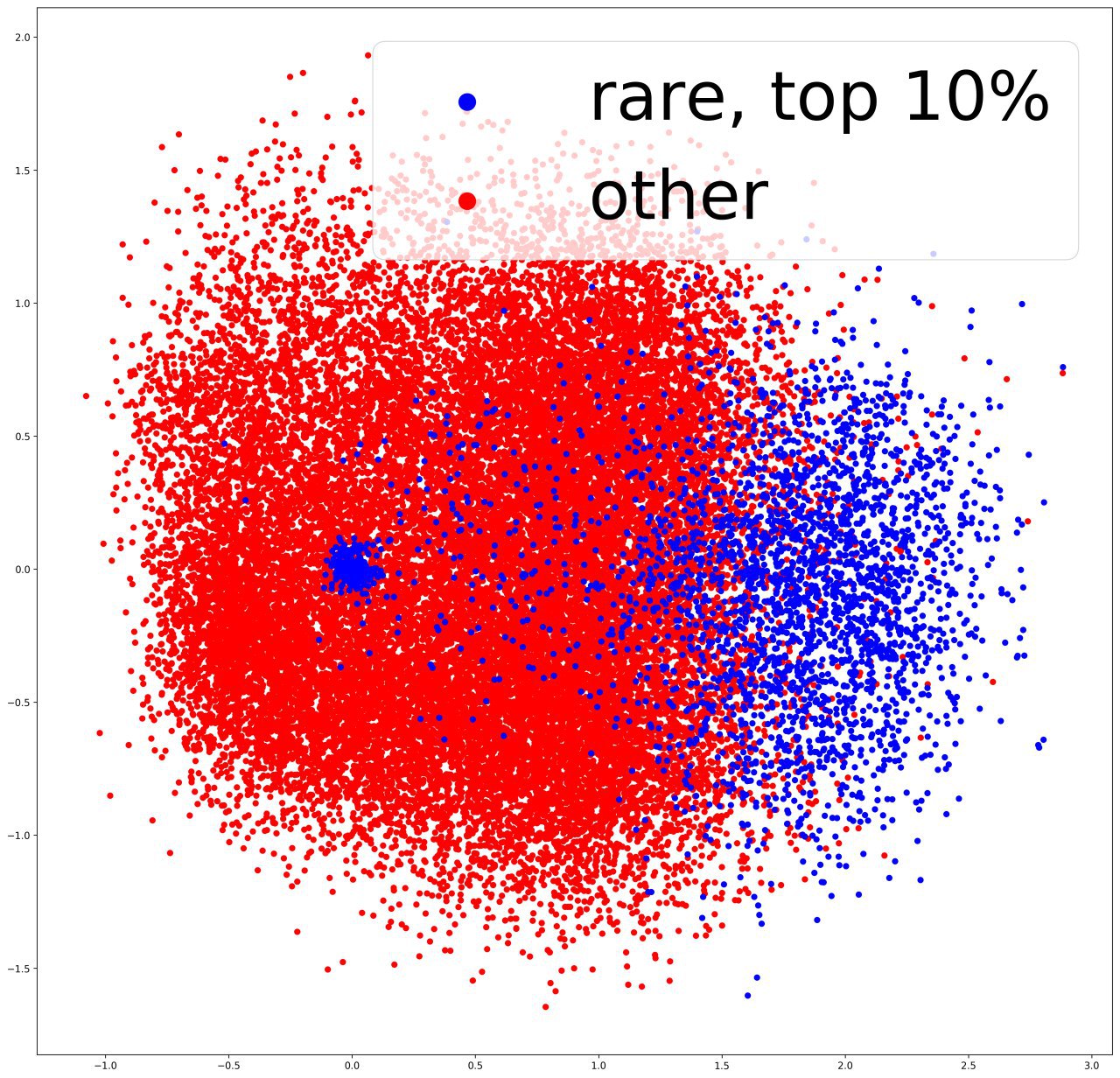}
        \caption{BPE}
    \end{subfigure}
    \quad
    \begin{subfigure}[b]{0.22\textwidth}
       \includegraphics[width=\textwidth]{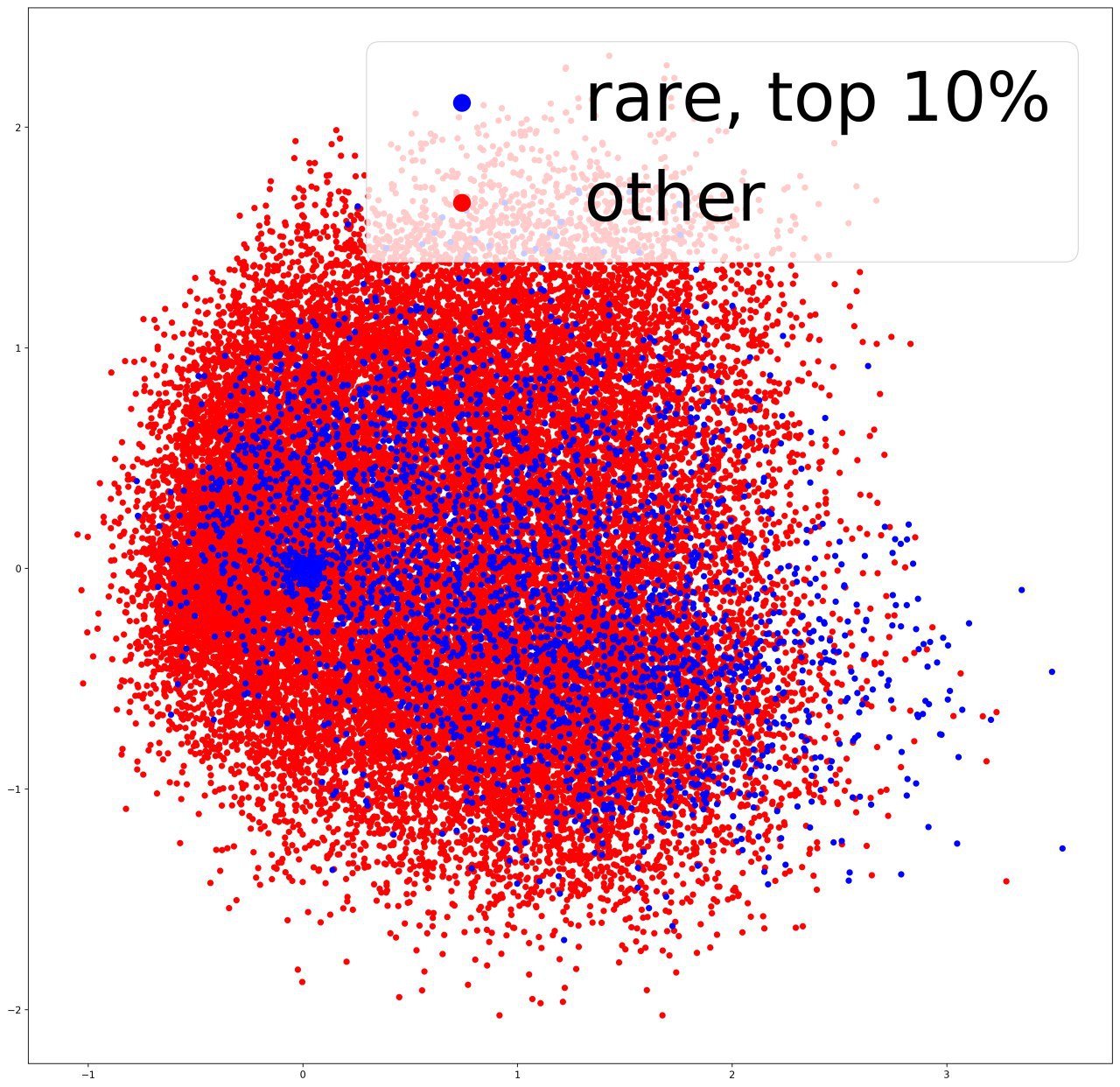}
        \caption{\textit{BPE-dropout}}
    \end{subfigure}
    \caption{Visualization of source embeddings. Models trained on WMT14 En-Fr (4m).}
    \label{fig:rare_embs}
\end{figure}

This also relates to the study by \citet{gong2018frage}. For several tasks, they analyze the embedding space learned by a model. The authors find that while a popular token usually has semantically related neighbors, a rare word usually does not: a vast majority of closest neighbors of rare words are rare words. To confirm this, we reduce dimensionality of embeddings by SVD and visualize~(Figure~\ref{fig:rare_embs}). For the model trained with BPE, rare tokens are in general separated from the rest; for the model trained with \textit{BPE-dropout}, this is not the case. While to alleviate this issue \citet{gong2018frage} propose to use adversarial training for embedding layers, we showed that a trained with \textit{BPE-dropout} model does not have this problem.

\subsection{Robustness to misspelled input}
\label{sect:misspel}

\newcommand{\red}[1]{\textcolor{red}{#1}}

\begin{table}[t!]
\centering
\begin{tabular}{lcccc}
\toprule
 & source & BPE & \textit{BPE-dropout} & diff\\

\midrule
\multicolumn{5}{l}{\!\!\!\bf En-De}\\
&original & 27.41 & \bf{28.01} & +0.6 \\
&misspelled & 24.45 & \bf{26.03} & +1.58\\
\midrule
\multicolumn{5}{l}{\!\!\!\bf De-En}\\
&\!\!\!\!\!\!  original & 32.69 & \bf{34.19} & +1.5\\
&\!\!\!\!\!\!  misspelled & 29.71 & \bf{32.03} & +2.32\\
\midrule
\multicolumn{5}{l}{\!\!\!\bf En-Fr (4m)}\\
&\!\!\!\!\!\!  original & 33.38 & \bf{33.85} & +0.47\\
&\!\!\!\!\!\!  misspelled & 30.30 & \bf{32.13} & +1.83 \\
\midrule
\multicolumn{5}{l}{\!\!\!\bf En-Fr (16m)}\\
&\!\!\!\!\!\!  original & 34.37 & \bf{34.82} & +0.45\\
&\!\!\!\!\!\!  misspelled & 31.23 & \bf{32.94} & +1.71 \\
\bottomrule
\end{tabular}
\caption{BLEU scores for models trained on WMT14 dataset evaluated given the original and misspelled source. For En-Fr trained on 16m sentence pairs, \textit{BPE-dropout} was used only on the source side (Section~\ref{sect:drop_on_one_side}).}
\label{tab:misspell}

\end{table}

Models trained with \textit{BPE-dropout} better learn compositionality of words and the meaning of subwords, which suggests that these models have to be more robust to noise. We verify this by measuring the translation quality of models on a test set augmented with synthetic misspellings.
We augment the source side of a test set by modifying each word with the probability of $10\%$ by applying one of the predefined operations. The operations we consider are (1) removal of one character from a word, (2) insertion of a random character into a word, (3) substitution of a character in a word with a random one. This augmentation produces words with the edit distance of 1 from the unmodified words. Edit distance is commonly used to model misspellings \cite{misspell2,misspell1,pinter-etal-2017-mimicking}.

Table~\ref{tab:misspell} shows the translation quality of the models trained on WMT 14 dataset when given the original source and augmented with misspellings. We deliberately chose large datasets, where improvements from using \textit{BPE-dropout} are smaller. We can see that while for the original test sets the improvements from using \textit{BPE-dropout} are usually modest, for misspelled test set the improvements are a lot larger: 1.6-2.3~BLEU. This is especially interesting since models have not been exposed to misspellings during training. Therefore, even for large datasets using \textit{BPE-dropout} can result in substantially better quality for practical applications where input is likely to be noisy.

\section{Related work}

Closest to our work in motivation is the work by \citet{sentencepiece}, who introduced the subword regularization framework multiple segmentation candidates 
and a new segmentation algorithm. 
Other segmentation  algorithms include \citet{creutz2006morfessor}, \citet{schuster2012japanese}, \citet{chitnis-denero-2015-variable}, \citet{kunchukuttan-bhattacharyya-2016-orthographic}, \citet{wu2018finding}, \citet{banerjee-bhattacharyya-2018-meaningless}.

Regularization techniques are widely used for training deep neural networks. Among regularizations applied to a network weights the most popular are Dropout~\cite{srivastava2014dropout} and $L_2$ regularization. Data augmentation techniques in natural language processing include dropping  tokens at random positions or swapping tokens at close positions~\cite{iyyer-etal-2015-deep,artetxe-etal-2018-unsupervised,lample2018unsupervised}, replacing tokens at random positions with a placeholder token~\cite{xie2017datanoising}, replacing tokens at random positions with a token sampled from some distribution (e.g., based on token frequency or a language model)~\cite{fadaee-etal-2017-data,xie2017datanoising,kobayashi-2018-contextual}. While \textit{BPE-dropout} can be thought of as a regularization, our motivation is not to make a model robust by injecting noise. By exposing a model to different segmentations, we want to teach it to better understand the composition of words as well as subwords, and make it more flexible in the choice of segmentation during inference.

Several works study how translation quality depends on a level of granularity of a segmentation~\cite{revisiting, char_preferred, lowresBPE}. 
\citet{revisiting} show that trained long enough character-level models tend to have better quality, but it comes with the increase of computational cost for both training and inference. \citet{char_preferred} find that, given flexibility in choosing segmentation level, the model prefers to operate on (almost) character level. \citet{lowresBPE} explore the effect of BPE vocabulary size and find that it is better to use small vocabulary for low-resource setting and large vocabulary for a high-resource setting. Following these observations, in our experiments we use different vocabulary size depending on a dataset size to ensure the strongest baselines.

\section{Conclusions} 

We introduce \textit{BPE-dropout} -- simple and effective subword regularization, which operates within the standard BPE framework. The only difference from BPE is how a word is segmented during model training: \textit{BPE-dropout} randomly drops some merges from the BPE merge table, which results in different segmentations for the same word. Models trained with \textit{BPE-dropout} (1)~outperform BPE and the previous subword regularization on a wide range of translation tasks, (2)~have better quality of learned embeddings, (3)~are more robust to noisy input. Future research directions include adaptive dropout rates for different merges and an in-depth analysis of other pathologies in learned token embeddings for different segmentations.

\section*{Acknowledgments} 
We thank anonymous reviewers for the helpful feedback, Rico Sennrich for valuable comments on the first version of this paper, and Yandex Machine Translation team for  discussions and inspiration.

\bibliography{acl2019}

\begin{thebibliography}{30}
\expandafter\ifx\csname natexlab\endcsname\relax\def\natexlab#1{#1}\fi

\bibitem[{Ahmad and Kondrak(2005)}]{misspell1}
Farooq Ahmad and Grzegorz Kondrak. 2005.
\newblock \href {https://www.aclweb.org/anthology/H05-1120} {Learning a
  spelling error model from search query logs}.
\newblock In \emph{Proceedings of Human Language Technology Conference and
  Conference on Empirical Methods in Natural Language Processing}, pages
  955--962, Vancouver, British Columbia, Canada. Association for Computational
  Linguistics.

\bibitem[{Artetxe et~al.(2018)Artetxe, Labaka, and
  Agirre}]{artetxe-etal-2018-unsupervised}
Mikel Artetxe, Gorka Labaka, and Eneko Agirre. 2018.
\newblock \href {https://doi.org/10.18653/v1/D18-1399} {Unsupervised
  statistical machine translation}.
\newblock In \emph{Proceedings of the 2018 Conference on Empirical Methods in
  Natural Language Processing}, pages 3632--3642, Brussels, Belgium.
  Association for Computational Linguistics.

\bibitem[{Banerjee and
  Bhattacharyya(2018)}]{banerjee-bhattacharyya-2018-meaningless}
Tamali Banerjee and Pushpak Bhattacharyya. 2018.
\newblock \href {https://doi.org/10.18653/v1/W18-1207} {Meaningless yet
  meaningful: Morphology grounded subword-level {NMT}}.
\newblock In \emph{Proceedings of the Second Workshop on Subword/Character
  {LE}vel Models}, pages 55--60, New Orleans. Association for Computational
  Linguistics.

\bibitem[{Barrault et~al.(2019)Barrault, Bojar, Costa-juss{\`a}, Federmann,
  Fishel, Graham, Haddow, Huck, Koehn, Malmasi, Monz, M{\"u}ller, Pal, Post,
  and Zampieri}]{barrault-etal-2019-findings}
Lo{\"\i}c Barrault, Ond{\v{r}}ej Bojar, Marta~R. Costa-juss{\`a}, Christian
  Federmann, Mark Fishel, Yvette Graham, Barry Haddow, Matthias Huck, Philipp
  Koehn, Shervin Malmasi, Christof Monz, Mathias M{\"u}ller, Santanu Pal, Matt
  Post, and Marcos Zampieri. 2019.
\newblock \href {https://doi.org/10.18653/v1/W19-5301} {Findings of the 2019
  conference on machine translation ({WMT}19)}.
\newblock In \emph{Proceedings of the Fourth Conference on Machine Translation
  (Volume 2: Shared Task Papers, Day 1)}, pages 1--61, Florence, Italy.
  Association for Computational Linguistics.

\bibitem[{Bojar et~al.(2018)Bojar, Federmann, Fishel, Graham, Haddow, Koehn,
  and Monz}]{bojar-etal-2018-findings}
Ond{\v{r}}ej Bojar, Christian Federmann, Mark Fishel, Yvette Graham, Barry
  Haddow, Philipp Koehn, and Christof Monz. 2018.
\newblock \href {https://doi.org/10.18653/v1/W18-6401} {Findings of the 2018
  conference on machine translation ({WMT}18)}.
\newblock In \emph{Proceedings of the Third Conference on Machine Translation:
  Shared Task Papers}, pages 272--303, Belgium, Brussels. Association for
  Computational Linguistics.

\bibitem[{Brill and Moore(2000)}]{misspell2}
Eric Brill and Robert~C. Moore. 2000.
\newblock \href {https://doi.org/10.3115/1075218.1075255} {An improved error
  model for noisy channel spelling correction}.
\newblock In \emph{Proceedings of the 38th Annual Meeting of the Association
  for Computational Linguistics}, pages 286--293, Hong Kong. Association for
  Computational Linguistics.

\bibitem[{Cherry et~al.(2018)Cherry, Foster, Bapna, Firat, and
  Macherey}]{revisiting}
Colin Cherry, George Foster, Ankur Bapna, Orhan Firat, and Wolfgang Macherey.
  2018.
\newblock \href {https://doi.org/10.18653/v1/D18-1461} {Revisiting
  character-based neural machine translation with capacity and compression}.
\newblock In \emph{Proceedings of the 2018 Conference on Empirical Methods in
  Natural Language Processing}, pages 4295--4305, Brussels, Belgium.
  Association for Computational Linguistics.

\bibitem[{Chitnis and DeNero(2015)}]{chitnis-denero-2015-variable}
Rohan Chitnis and John DeNero. 2015.
\newblock \href {https://doi.org/10.18653/v1/D15-1249} {Variable-length word
  encodings for neural translation models}.
\newblock In \emph{Proceedings of the 2015 Conference on Empirical Methods in
  Natural Language Processing}, pages 2088--2093, Lisbon, Portugal. Association
  for Computational Linguistics.

\bibitem[{Creutz and Lagus(2006)}]{creutz2006morfessor}
Mathias Creutz and Krista Lagus. 2006.
\newblock \href
  {http://citeseerx.ist.psu.edu/viewdoc/download?doi=10.1.1.78.1179&rep=rep1&type=pdf}
  {Morfessor in the morpho challenge}.
\newblock In \emph{Proceedings of the PASCAL Challenge Workshop on Unsupervised
  Segmentation of Words into Morphemes}, pages 12--17. Citeseer.

\bibitem[{Ding et~al.(2019)Ding, Renduchintala, and Duh}]{lowresBPE}
Shuoyang Ding, Adithya Renduchintala, and Kevin Duh. 2019.
\newblock \href {https://www.aclweb.org/anthology/W19-6620.pdf} {A call for
  prudent choice of subword merge operations in neural machine translation}.
\newblock In \emph{Proceedings of Machine Translation Summit XVII Volume 1:
  Research Track}, pages 204--213, Dublin, Ireland. European Association for
  Machine Translation.

\bibitem[{Fadaee et~al.(2017)Fadaee, Bisazza, and Monz}]{fadaee-etal-2017-data}
Marzieh Fadaee, Arianna Bisazza, and Christof Monz. 2017.
\newblock \href {https://doi.org/10.18653/v1/P17-2090} {Data augmentation for
  low-resource neural machine translation}.
\newblock In \emph{Proceedings of the 55th Annual Meeting of the Association
  for Computational Linguistics (Volume 2: Short Papers)}, pages 567--573,
  Vancouver, Canada. Association for Computational Linguistics.

\bibitem[{Gong et~al.(2018)Gong, He, Tan, Qin, Wang, and Liu}]{gong2018frage}
Chengyue Gong, Di~He, Xu~Tan, Tao Qin, Liwei Wang, and Tie-Yan Liu. 2018.
\newblock \href
  {https://papers.nips.cc/paper/7408-frage-frequency-agnostic-word-representation.pdf}
  {Frage: Frequency-agnostic word representation}.
\newblock In \emph{Advances in neural information processing systems}, pages
  1334--1345.

\bibitem[{Iyyer et~al.(2015)Iyyer, Manjunatha, Boyd-Graber, and
  Daum{\'e}~III}]{iyyer-etal-2015-deep}
Mohit Iyyer, Varun Manjunatha, Jordan Boyd-Graber, and Hal Daum{\'e}~III. 2015.
\newblock \href {https://doi.org/10.3115/v1/P15-1162} {Deep unordered
  composition rivals syntactic methods for text classification}.
\newblock In \emph{Proceedings of the 53rd Annual Meeting of the Association
  for Computational Linguistics and the 7th International Joint Conference on
  Natural Language Processing (Volume 1: Long Papers)}, pages 1681--1691,
  Beijing, China. Association for Computational Linguistics.

\bibitem[{Kobayashi(2018)}]{kobayashi-2018-contextual}
Sosuke Kobayashi. 2018.
\newblock \href {https://doi.org/10.18653/v1/N18-2072} {Contextual
  augmentation: Data augmentation by words with paradigmatic relations}.
\newblock In \emph{Proceedings of the 2018 Conference of the North {A}merican
  Chapter of the Association for Computational Linguistics: Human Language
  Technologies, Volume 2 (Short Papers)}, pages 452--457, New Orleans,
  Louisiana. Association for Computational Linguistics.

\bibitem[{Koehn(2004)}]{koehn2004statistical}
Philipp Koehn. 2004.
\newblock \href {https://www.aclweb.org/anthology/W04-3250} {Statistical
  significance tests for machine translation evaluation}.
\newblock In \emph{Proceedings of the 2004 Conference on Empirical Methods in
  Natural Language Processing}.

\bibitem[{Kreutzer and Sokolov(2018)}]{char_preferred}
Julia Kreutzer and Artem Sokolov. 2018.
\newblock \href
  {https://www.cl.uni-heidelberg.de/~sokolov/pubs/kreutzer18learning.pdf}
  {Learning to segment inputs for nmt favors character-level processing}.
\newblock In \emph{Proceedings of the 15th International Workshop on Spoken
  Language Translation}.

\bibitem[{Kudo(2018)}]{sentencepiece}
Taku Kudo. 2018.
\newblock \href {https://www.aclweb.org/anthology/P18-1007} {Subword
  regularization: Improving neural network translation models with multiple
  subword candidates}.
\newblock In \emph{Proceedings of the 56th Annual Meeting of the Association
  for Computational Linguistics (Volume 1: Long Papers)}, pages 66--75,
  Melbourne, Australia. Association for Computational Linguistics.

\bibitem[{Kunchukuttan and
  Bhattacharyya(2016)}]{kunchukuttan-bhattacharyya-2016-orthographic}
Anoop Kunchukuttan and Pushpak Bhattacharyya. 2016.
\newblock \href {https://doi.org/10.18653/v1/D16-1196} {Orthographic syllable
  as basic unit for {SMT} between related languages}.
\newblock In \emph{Proceedings of the 2016 Conference on Empirical Methods in
  Natural Language Processing}, pages 1912--1917, Austin, Texas. Association
  for Computational Linguistics.

\bibitem[{Lample et~al.(2018)Lample, Conneau, Denoyer, and
  Ranzato}]{lample2018unsupervised}
Guillaume Lample, Alexis Conneau, Ludovic Denoyer, and Marc'Aurelio Ranzato.
  2018.
\newblock \href {https://openreview.net/forum?id=rkYTTf-AZ} {Unsupervised
  machine translation using monolingual corpora only}.
\newblock In \emph{International Conference on Learning Representations}.

\bibitem[{Ott et~al.(2018)Ott, Auli, Grangier, and Ranzato}]{ott2018analyzing}
Myle Ott, Michael Auli, David Grangier, and Marc’Aurelio Ranzato. 2018.
\newblock Analyzing uncertainty in neural machine translation.
\newblock In \emph{International Conference on Machine Learning}, pages
  3956--3965.

\bibitem[{Papineni et~al.(2002)Papineni, Roukos, Ward, and Zhu}]{BLEU}
Kishore Papineni, Salim Roukos, Todd Ward, and Wei-Jing Zhu. 2002.
\newblock \href {https://doi.org/10.3115/1073083.1073135} {{B}leu: a method for
  automatic evaluation of machine translation}.
\newblock In \emph{Proceedings of the 40th Annual Meeting of the Association
  for Computational Linguistics}, pages 311--318, Philadelphia, Pennsylvania,
  USA. Association for Computational Linguistics.

\bibitem[{Pinter et~al.(2017)Pinter, Guthrie, and
  Eisenstein}]{pinter-etal-2017-mimicking}
Yuval Pinter, Robert Guthrie, and Jacob Eisenstein. 2017.
\newblock \href {https://doi.org/10.18653/v1/D17-1010} {Mimicking word
  embeddings using subword {RNN}s}.
\newblock In \emph{Proceedings of the 2017 Conference on Empirical Methods in
  Natural Language Processing}, pages 102--112, Copenhagen, Denmark.
  Association for Computational Linguistics.

\bibitem[{Post(2018)}]{sacrebleu}
Matt Post. 2018.
\newblock \href {https://www.aclweb.org/anthology/W18-6319} {A call for clarity
  in reporting {BLEU} scores}.
\newblock In \emph{Proceedings of the Third Conference on Machine Translation:
  Research Papers}, pages 186--191, Belgium, Brussels. Association for
  Computational Linguistics.

\bibitem[{Schuster and Nakajima(2012)}]{schuster2012japanese}
Mike Schuster and Kaisuke Nakajima. 2012.
\newblock \href {https://ieeexplore.ieee.org/document/6289079} {Japanese and
  korean voice search}.
\newblock In \emph{2012 IEEE International Conference on Acoustics, Speech and
  Signal Processing (ICASSP)}, pages 5149--5152. IEEE.

\bibitem[{Sennrich et~al.(2016)Sennrich, Haddow, and
  Birch}]{sennrich-etal-2016-neural}
Rico Sennrich, Barry Haddow, and Alexandra Birch. 2016.
\newblock \href {https://doi.org/10.18653/v1/P16-1162} {Neural machine
  translation of rare words with subword units}.
\newblock In \emph{Proceedings of the 54th Annual Meeting of the Association
  for Computational Linguistics (Volume 1: Long Papers)}, pages 1715--1725,
  Berlin, Germany. Association for Computational Linguistics.

\bibitem[{Sennrich and Zhang(2019)}]{sennrich-zhang-2019-revisiting}
Rico Sennrich and Biao Zhang. 2019.
\newblock \href {https://doi.org/10.18653/v1/P19-1021} {Revisiting low-resource
  neural machine translation: A case study}.
\newblock In \emph{Proceedings of the 57th Annual Meeting of the Association
  for Computational Linguistics}, pages 211--221, Florence, Italy. Association
  for Computational Linguistics.

\bibitem[{Srivastava et~al.(2014)Srivastava, Hinton, Krizhevsky, Sutskever, and
  Salakhutdinov}]{srivastava2014dropout}
Nitish Srivastava, Geoffrey Hinton, Alex Krizhevsky, Ilya Sutskever, and Ruslan
  Salakhutdinov. 2014.
\newblock \href
  {http://www.jmlr.org/papers/volume15/srivastava14a/srivastava14a.pdf}
  {Dropout: a simple way to prevent neural networks from overfitting}.
\newblock \emph{The journal of machine learning research}, 15(1):1929--1958.

\bibitem[{Vaswani et~al.(2017)Vaswani, Shazeer, Parmar, Uszkoreit, Jones,
  Gomez, Kaiser, and Polosukhin}]{transformer}
Ashish Vaswani, Noam Shazeer, Niki Parmar, Jakob Uszkoreit, Llion Jones,
  Aidan~N Gomez, Lukasz Kaiser, and Illia Polosukhin. 2017.
\newblock \href
  {http://papers.nips.cc/paper/7181-attention-is-all-you-need.pdf} {Attention
  is all you need}.
\newblock In \emph{NeurIPS}, Los Angeles.

\bibitem[{Wu and Zhao(2018)}]{wu2018finding}
Yingting Wu and Hai Zhao. 2018.
\newblock \href {https://arxiv.org/pdf/1807.09639.pdf} {Finding better subword
  segmentation for neural machine translation}.
\newblock In \emph{Chinese Computational Linguistics and Natural Language
  Processing Based on Naturally Annotated Big Data}, pages 53--64. Springer.

\bibitem[{Xie et~al.(2017)Xie, Wang, Li, Lévy, Nie, Jurafsky, and
  Ng}]{xie2017datanoising}
Ziang Xie, Sida~I. Wang, Jiwei Li, Daniel Lévy, Aiming Nie, Dan Jurafsky, and
  Andrew~Y. Ng. 2017.
\newblock \href {https://openreview.net/forum?id=H1VyHY9gg} {Data noising as
  smoothing in neural network language models}.
\newblock In \emph{International Conference on Learning Representations}.

\end{thebibliography}
\bibliographystyle{acl_natbib}

\appendix

\section{Training time}

Table~\ref{tab:num_steps_main} shows
number of training batches for the experiments in Section~\ref{sect:main_results} (Table~\ref{tab:main}), Table~\ref{tab:num_steps_one_side}~--- for the experiments in Section~\ref{sect:drop_on_one_side} (Table~\ref{tab:one_side}).

\begin{table}[h!]
\centering
\begin{tabular}{lcccc}
\toprule
 & & BPE  & \citet{sentencepiece} & \textit{BPE-dropout}\\

\midrule
\multicolumn{5}{l}{\!\!\!\bf IWSLT15}\\
&\!\!\!\!\!\! En-Vi & 23 & 26 & 36 \\
&\!\!\!\!\!\! Vi-En & 23 & 29 & 33 \\
&\!\!\!\!\!\! En-Zh & 30 & 29  & 43 \\
&\!\!\!\!\!\! Zh-En & 39 & 51 & 100 \\
\midrule
\multicolumn{5}{l}{\!\!\!\bf IWSLT17}\\
&\!\!\!\!\!\! En-Fr & 36 & 45 & 60 \\
&\!\!\!\!\!\! Fr-En & 32 & 46 & 85 \\
&\!\!\!\!\!\! En-Ar & 30 & 60 & 62 \\
&\!\!\!\!\!\! Ar-En & 41 & 51 & 59 \\
\midrule
\multicolumn{5}{l}{\!\!\!\bf WMT14}\\
&\!\!\!\!\!\! En-De & 468 & 450 & 501  \\
&\!\!\!\!\!\! De-En & 447 & 442 & 525\\
\midrule
\multicolumn{5}{l}{\!\!\!\bf ASPEC}\\
&\!\!\!\!\!\! En-Ja & 280 & 165 & 462\\
&\!\!\!\!\!\! Ja-En & 239 & 144 & 576 \\

\bottomrule
\end{tabular}
\caption{Number of thousands of training batches for the experiments from Table~\ref{tab:main}.}
\label{tab:num_steps_main}
\end{table}

\begin{table}[h!]
\centering
\begin{tabular}{lcccc}
\toprule
 & BPE & \multicolumn{3}{c}{\textit{BPE-dropout}}\\
 & & src-only  & dst-only & both\\

\midrule
250k & 47 & 53 & 53 & 85 \\
500k & 160 & 210 & 250 & 320 \\
1m & 30 & 114 & 67 & 180 \\
4m & 100 & 321 & 180 & 600 \\
16m & 345 & 345 & - & 400 \\
\bottomrule
\end{tabular}
\caption{Number of thousands of training batches for the experiments from Table~\ref{tab:one_side}. Note that we use batch size 4k tokens for small corpora (250k and 500k) and 32k tokens for large corpora (1m, 4m and 16m).}
\label{tab:num_steps_one_side}
\end{table}

\section{Additional experiments}

In the main text, all models were trained (and evaluated) on lowercased data. Here we provide results of the models trained and evaluated without lower case (Table~\ref{tab:main_no_lowersace}).

\begin{table}[h!]
\centering
\begin{tabular}{lccc}
\toprule
 & & BPE  & \textit{BPE-dropout}\\

\midrule
\multicolumn{4}{l}{\!\!\!\bf IWSLT15}\\
& En-Vi & 31{.}44 & \bf{32{.}70} \\
& Vi-En & 32{.}19 & \bf{33{.}22} \\
\midrule
\multicolumn{4}{l}{\!\!\!\bf IWSLT17}\\
& En-Fr & 38{.}79 &	\bf{39{.}83} \\
& Fr-En & 38{.}06 &	\bf{38{.}60} \\
& En-Ar & 14.30 & \bf{15.20} \\
& Ar-En & 31.56 & \bf{33.00} \\

\bottomrule
\end{tabular}
\caption{BLEU scores. Bold indicates the best score; differences with the baselines are statistically significant (with $p$-value of 0.05). (Statistical significance is computed via bootstrapping~\cite{koehn2004statistical}.)}
\label{tab:main_no_lowersace}
\end{table}

\end{document}